# Secure short-term load forecasting for smart grids with transformer-based federated learning

J. Sievers*, and T. Blank*
*Karlsruhe Institute of Technology (KIT), Institute for Data Processing and Electronics (IPE),
Hermann-von-Helmholtz-Platz 1, 76344 Eggenstein-Leopoldshafen, (*Germany*)

*Abstract* — Electricity load forecasting is an essential task within smart grids to assist demand and supply balance. While advanced deep learning models require large amounts of high-resolution data for accurate short-term load predictions, fine-grained load profiles can expose users' electricity consumption behaviors, which raises privacy and security concerns. One solution to improve data privacy is federated learning, where models are trained locally on private data, and only the trained model parameters are merged and updated on a global server. Therefore, this paper presents a novel transformer-based deep learning approach with federated learning for short-term electricity load prediction. To evaluate our results, we benchmark our federated learning architecture against central and local learning and compare the performance of our model to long short-term memory models and convolutional neural networks. Our simulations are based on a dataset from a German university campus and show that transformer-based forecasting is a promising alternative to state-of-the-art models within federated learning.

*Index Terms*-- federated learning, load forecasting, smart grid, transformer

## I. Introduction

The requirements for a reliable, resilient, and secure energy management architecture are growing [1]. While the energy crisis and climate change have accelerated the installation of photovoltaic and wind turbines, the electricity supply is increasingly volatile and uncertain. Therefore, electric load forecasting, especially short-term load forecasting (STLF), is essential for planning and operating smart grids. Accurate STLF supports multiple disciplines, including power dispatching, intra-day generation planning, and peak load shaving [1]. Over the last decades, STLF models have improved significantly, and nowadays, long short-term memory (LSTM) models and convolutional neural networks (CNNs) are widely used in time series forecasting, as they can capture complex and non-linear patterns [2]. A high volume of fine-grained data is essential for accurate prediction models. Therefore, smart grids and industrial sites install advanced metering infrastructure to monitor real-time energy consumption [3]. For data processing and model training, three fundamental architectures exist, namely central, local, and federated learning (FL).

In central learning, the data is uploaded to a central server, and one forecasting model is used to predict all load profiles. However, regulatory authorities and users have raised many privacy concerns [4]. As research shows, attackers can use high-resolution electricity consumption data to reveal customers' habits [4], location [5], or customers' absences for break-ins [2], leading users in some countries to refuse smart meter installation. Previous attempts to improve data security, like data aggregation, are unsuitable for precise STLF models, which need fine-grained input data [6].

Another alternative is local learning, where each party stores its data locally and builds its individual STLF model. However, this prevents the forecasting models from potentially benefiting from peers' data and thus limits scalability and transfer learning [7].

FL has recently been proposed as a solution introducing a distributed machine learning approach to improve the data security, privacy, latency, and bandwidth of the underlying communication network [8]. Here, models are trained locally on private data, and only the model parameters are shared and updated on a global server [9].

### A. Related work

FL was introduced by Vaswani et al. [10] in 2017 as a novel technique to share prediction models from different mobile phones collaboratively. Within the last years, some researchers have adopted FL within the energy domain, including energy control [1, 3, 11], non-intrusive load monitoring [12–14], and energy theft detection in smart grids [15]. In particular, the implementation of FL for STLF has been investigated in some publications, as seen in Table 1.

Within the STLF domain, authors focus on electricity forecasting [16] or multi-energy predictions [17] and analyze attack scenarios [18] or privacy-enhancing measures like noise adding [2]. Since highly heterogenous data from different facilities can reduce FL performance, several authors investigate the clustering of similar datasets to improve prediction accuracy. Here clustering techniques like hierarchical clustering [19], k-means clustering [20], and socioeconomic clustering [4] are analyzed.

This work was funded by the German Research Foundation (DFG) as part of the Research Training Group 2153: 'Energy Status Data - Informatics Methods for its Collection, Analysis, and Exploitation'.

TABLE 1: CONCEPT MATRIX ON THE LITERATURE OF FL-BASED STLF

| Ref. | Year | Focus | LSTM | CNN | Transformer |
|---|---|---|---|---|---|
| [2] | 2021 | Increasing privacy by adding noise | ✓ | | |
| [20] | 2022 | Effect of clustering on forecast accuracy | ✓ | | |
| [5] | 2020 | Effect on networking load gain | ✓ | | |
| [21] | 2021 | Effect of clustered aggregation | ✓ | | |
| [19] | 2022 | Effect of different clustering variants | ✓ | | |
| [18] | 2022 | Poisoning attack on FL architecture | ✓ | | |
| [4] | 2021 | Socioeconomic clustering for FL | ✓ | | |
| [14] | 2021 | Effect of different clustering variants | ✓ | | |
| [16] | 2022 | FL for STLF | ✓ | | |
| [22] | 2022 | Comparing LSTM with basic models | ✓ | | |
| [23] | 2022 | Adaptative federated transfer learning | | ✓ | |
| [17] | 2022 | Multi-energy FL forecasting | | ✓ | |
| **This paper** | **2023** | **FL for STLF with CNNs, LSTMs, and transformers** | ✓ | ✓ | ✓ |

Over the last decades, STLF models have improved significantly, and LSTM models have outperformed traditional recurrent neural networks (RNNs) for load forecasting [16]. Therefore, within the analyzed literature, most publications apply LSTM models for forecasting [2, 4, 5, 14, 16, 18–22], while only a few consider CNNs or hybrid LSTM-CNNs [17, 23]. However, LSTM models are challenging to parallelize due to their sequential nature, limiting their scalability. In response to high computational cost, the transformer-based architecture has been developed in the natural language processing domain [10], outperforming LSTM models and CNNs while also allowing parallelization [24]. Since then, the application range of transformers has been extended to various domains, including computer vision [25], reinforcement learning [26], and audio [27].

L'Heureux et al. [24] first present a transformer-based architecture for load forecasting by modifying the architecture, adding N-space transformation, and proposing a procedure for contextual feature handling. Results show that the transformer-based forecasting model outperforms other state-of-the-art methods in accuracy and scalability.

*B. Paper contribution and organization*

However, to the best of our knowledge, no publications exist that propose a transformer-based FL architecture for STLF. Therefore, we focus on analyzing the forecasting performance of a novel transformer-based prediction model compared to selected state-of-the-art deep learning methods. Consequently, our main contributions are:

- First, we present a novel transformer-based FL architecture for STLF to improve data privacy, model robustness, and scalability while enhancing forecasting performance.
- Second, we compare our transformer-based model with state-of-the-art LSTM models and CNNs.
- Third, we implement each of the three models for a local, central, and FL architecture to evaluate our prediction results extensively.
- Fourth, we analyze the effect of two forecasting horizons (12 and 24 hours) and different calendarial and weather features for each model.
- Fifth, we demonstrate the effect of limited data on forecasting accuracy within local and FL.

The results show that our transformer-based FL architecture can improve forecasting accuracy while providing high privacy and scalability. We organize the rest of the paper, as seen in FIGURE 1: Section 2 introduces our data analysis, while Section 3 provides an overview of our forecasting models. Section 4 outlines our training scenarios, Section 5 evaluates our forecasting performance, and Section 6 presents our conclusion.

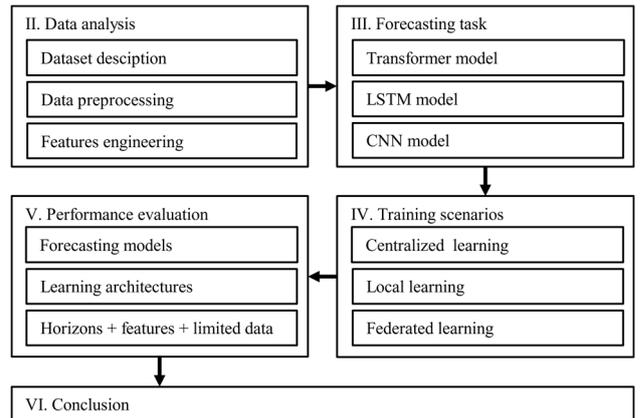

FIGURE 1: GRAPHICAL SUMMARY OF THE PAPER

II. DATA ANALYSIS

This section briefly describes our dataset, preprocessing techniques and explains our feature engineering methods, including Pearson correlation, min-max scaling, sine-cosine transformation, and Fast Fourier Transformation (FFT). We select the meter readings from a German university campus of the "Karlsruhe Institute of Technology" as a suitable dataset, which includes over 2000 smart meter readings from residential buildings and industrial sites between 2019 and 2021 in hourly resolution. The measured values are indicated in kW and can be seen in FIGURE 2. Within the selected datasets are administrative buildings, workshops, and production facilities. For performance reasons, we limit our simulations to a subset of 33 randomly chosen smart meter readings, hereafter referred to as clients. Next, we clean the data by filling in single missing values with the last non-null value. Further, we replace outliers with the median of the previous three load readings. We consider values as outliers that are (i) negative (no energy generation) or that (ii) deviate more than two standard deviations from the last value (load inertia on building level).

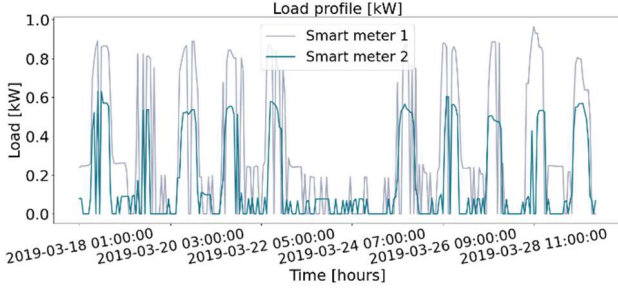

FIGURE 2: EXEMPLARY LOAD PROFILES FROM THE DATASET

In STLF, selecting the right features is crucial for building an accurate and effective forecasting model, as features provide necessary information about historical trends and patterns within the data. For feature engineering, we analyze weather-related and calendar features. To select promising weather features, we prepare a weather dataset from Meteostat [28] for our location and period. The dataset includes 14 features such as air temperature [°C], relative humidity [%], total precipitation [mm], average wind speed [km/h], sea-level air pressure [hPa], and others. Further, we calculate the apparent temperature according to Savi et al. [4], as they perform well with this feature. To select a final subset of weather-related features, we apply the Pearson correlation, which describes the linear relationship between two quantitative variables X and Y (1). The computed correlation values range from -1 to 1, indicating a strong negative or positive correlation.

$$r = \frac{Cov(X,Y)}{s_x * s_y} = \frac{n(\sum xy)-(\sum x)(\sum y)}{\sqrt{(n(\sum x^2-(\sum x)^2)(n\sum y^2-(\sum y)^2)}}, \quad (1)$$

where $Cov(X,Y)$ is the covariance and $s_i$ the standard deviation.

It is worth mentioning that not all variables necessarily have a linear relationship, and therefore, also Spearman correlation could have been used [29]. We then visualize the results in a heatmap, as seen in FIGURE 3, and select the two features with the highest correlation regarding the load: air temperature (temp, 0.18) and relative humidity (rhum, -0.21).

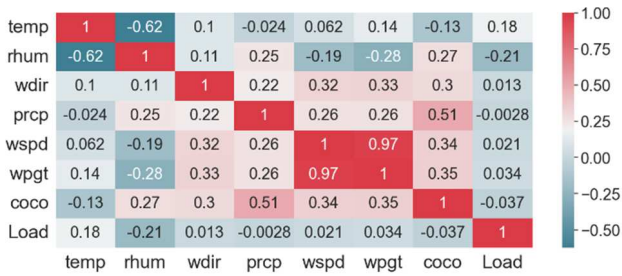

FIGURE 3: HEATMAP OF PEARSON CORRELATIONS

Afterward, we calculate the Pearson correlation for calendar features, including hour, weekday, month, quarter, and year. Here we choose the two features, hour and weekday, as they have the highest correlation. After feature selection, we map the cyclical variables hour of the day and weekday to the unit circle to allow periodicity in time. The mapping is essential, as 11 p.m. and 0 a.m. are consecutive times, whereas the numerical values 11 and 0 are not. For both features, we generate two sine and cosine features, as demonstrated in (2) [30]:

$$hour_{day} = \begin{cases} \sin(\frac{2\pi}{24} t) \\ \cos(\frac{2\pi}{24} t) \end{cases}, \quad (2)$$

where $t$ is the time value to be transformed.

Next, we use a min–max scaler (3) to scale all variables to have the same range from 0 to 1. Scaled transformation prevents attributes with larger scales from biasing the values of the objective functions, speeding up convergence [23].

$$x = \frac{x-x_{min}}{x_{max}-x_{min}}, \quad (3)$$

where $x$ is the variable value.

The forecasting models make predictions based on a window of consecutive data samples. To select a suitable window size, we apply the FFT on the load data to find the period of dominating seasonal components of the time series. FFT is a transformation function that converts the time series sequence $x_n$ to the frequency domain $x_k$. We calculate the Fourier coefficients (4) and plot the results in FIGURE 4.

$$x_k = \sum_{n=0}^{N-1} x_n e^{-i2\pi kn/N}, \quad (4)$$

where $x_k$ is the $k$th coefficient of the FFT and $X_n$ denotes the $n$th sample of the time series consisting of N samples.

According to the FFT, we use the last 24 hours as a sliding window for our forecasting models.

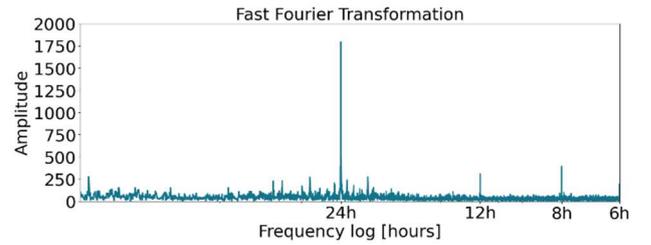

FIGURE 4: PLOTTED FOURIER COEFFICIENTS OF THE LOAD DATA

Finally, we split the dataset using 70% for training, 20% for validation, and 10% for testing to obtain our datasets for our forecasting models. It is worth mentioning that we maintain the continuity of the time series data by not shuffling the datasets.

III. FORECASTING TASK

We evaluate four forecasting tasks for each model, considering two forecasting horizons (12 and 24 hours) and two feature sets. In the first feature set, we select five features: energy consumption and the respective sine and cosine values of the hour and weekday. The second feature set consists of the last five features plus the air temperature and relative humidity. Therefore, we implement four sliding windows. Each with a look-back of 24 samples (one day), considering a look-ahead of sizes 12 and 24 and five and seven features. To evaluate our model accuracy, we select suitable evaluation metrics. FL models tend to predict the average of each dataset, hence offering only promising root mean squared errors (RMSE, (5)) and mean absolute errors (MAE, (6)). Therefore, we include also mean absolute percentage error (MAPE, (7)) [82]. Further, we measure the training time per epoch to evaluate the model's scalability and complexity.

$$RMSE = \sqrt{\frac{1}{n}\sum_{i=1}^{n}(y_i - x_i)^2}, \quad (5)$$

$$MAE = \frac{1}{n}\sum_{i=1}^{n}|y_i - x_i|, \quad (6)$$

$$MAPE = \frac{100}{n}\sum_{i=1}^{n}\left|\frac{x_i - y_i}{x_i}\right|, \quad (7)$$

where $y_i$ and $x_i$ present the predicted and real load value.

*A. Transformer Model*

The main structure of the transformer model consists of three components: the embedding, the encoder, and the decoder. The transformer embedding involves positional encoding and data preprocessing. In our simulation, we perform this step in advance during our data processing to use the same data for all three models. Together, the encoder-decoder of the transformer allows for the effective processing of time series data by leveraging self-attention mechanisms [31].

We choose the hyperparameters of our transformer model after careful manual tuning. Several architectures are tested, considering 1-6 encoder and decoder layers. Within the multi-head attention layers, we try different numbers of attention heads (2-4) with varying sizes (2-8). As seen in FIGURE 5, our best-performing transformer model consists of two encoder and two decoder layers.

The encoder input layer takes as input the window with a look-back of size 24 and passes it to the encoder. Within each encoder layer, the input tensor is fed into a multi-head attention layer with a dropout of 0.2. Next, we implement a residual connection (add layer) to connect the output of the multi-head attention layer with the input data before. The multi-head attention layers have 2 heads with a size of 4. The output is then passed to a dense layer, with a rectified linear unit (ReLU) activation function and another add layer and layer normalization. Finally, the output is forwarded into an LSTM layer with seven LSTM cells, where the hidden states are passed to the multi-head attention layer of the decoder. Within the decoder, each layer similarly consists of a multi-head attention layer, followed by add, dense, and layer normalization. The final outputs of the decoder are fed into a one-dimensional average pooling and a dense layer, with 12 or 24 neurons, depending on the forecasting horizon.

*B. LSTM Model*

An LSTM model consists of different memory cells which store information, along with input, forget, and output gates to control the information flow into and out of the cell [20]. In contrast to traditional RNNs, which can struggle to retain information over long time intervals due to the vanishing gradient problem, LSTM models can selectively remember or forget information over arbitrary time intervals [19].

To develop our final LSTM model, we carefully compare various architectures. Here we consider 1-10 LSTM layers with 2-256 LSTM cells and test the performance of adding 1-5 dense layers at the end. As seen in FIGURE 6, our final model consists of 6 LSTM layers with 32 LSTM cells each.

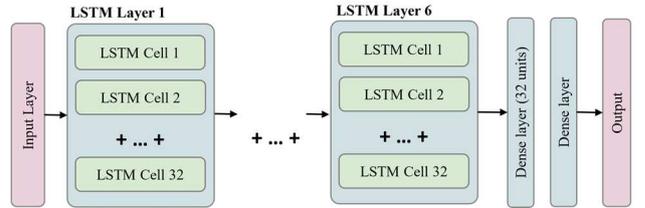

FIGURE 6: ARCHITECTURE OF OUR LSTM MODEL

The first layer takes 24 historical samples as input, passing them to the six consecutive LSTM layers with 32 LSTM cells each and hyperbolic tangent (tanh) activation functions. The last LSTM layer feeds the output into a dense layer with 32 units and a dropout of 0.2, followed by a final dense layer, with 12 or 24 neurons, depending on the forecasting horizon.

*C. CNN Model*

A CNN typically consists of a series of convolution and pooling layers for feature extraction followed by one or more fully connected layers to map the extracted features to the final output [32]. A convolutional layer operates in the time series domain by sliding a small kernel window across the time series data to extract features from the input tensor. The resulting feature map can be input for subsequent layers. A pooling layer provides downsampling and reduces the dimensionality of the input data by aggregating adjacent values into a single value using a predefined aggregation function [33].

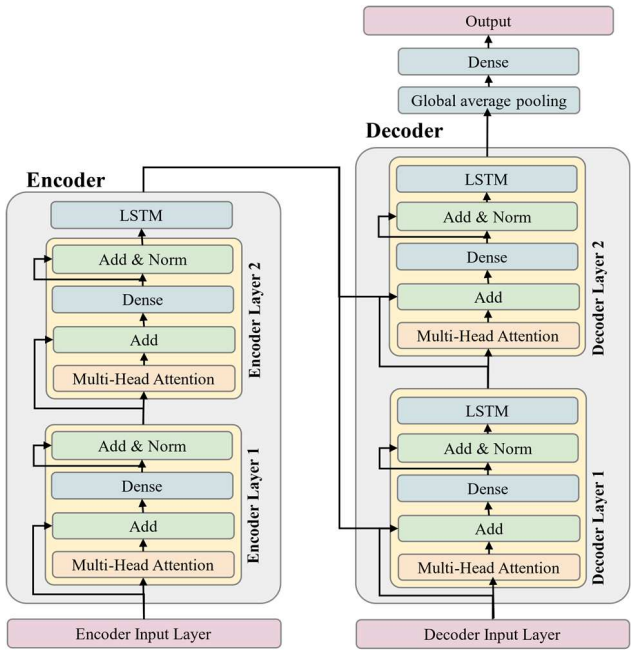

FIGURE 5: ARCHITECTURE OF OUR TRANSFORMER MODEL

We test different CNN architectures with varying convolutional layers (1-10), pooling layers (1-10), and additional batch-normalization layers. Further, we test various filter sizes (2-64) and convolutional widths (3-5). As seen in FIGURE 7, our final CNN model consists of 4 convolutional and batch-normalization layers, followed by an average pooling layer and one dense layer with 32 neurons and a 0.2 dropout. The convolutional layers have a convolution width of 3 and 32 filters.

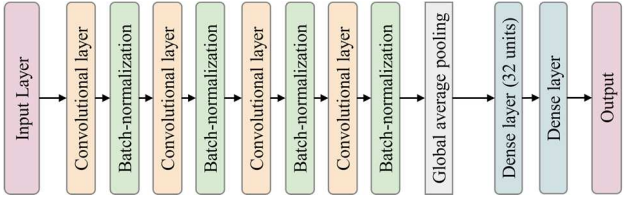

FIGURE 7: ARCHITECTURE OF OUR ONE-DIMENSIONAL CNN

## IV. TRAINING SCENARIOS

To evaluate the performance of our forecasting models, we introduce central and local learning as benchmarks. Within the three training architectures, we train our three models considering two forecasting horizons (12 and 24 hours) and the two feature sets, resulting in 36 training scenarios. All experiments are run in the TensorFlow 2 deep learning framework [34] on a simulation server using an Intel UHD Graphics 630 GPU with 16 GB memory attached to an Intel Core i9-9900K CPU at 4.6 GHz, with 8 kernels and 32 GB memory. The distributed training scenarios are all simulated on a single machine.

### A. Central Learning

In central learning, a non-distributed server merges all individual energy consumption datasets and performs data processing and model training. Here, a joint forecasting model is developed that has access to all datasets and thus can benefit from other users' data. However, the model must generalize the training data to provide high-quality predictions for the individual datasets. This approach is most commonly applied when data privacy is not a significant concern. Under the central learning scenario, we train our models for 100 epochs. Further, we consider early stopping based on the lowest error achieved on the validation set and the model checkpoint callback [19].

### B. Local Learning

All individual datasets remain private and unshared in the fully-private local learning architecture. Each client trains its unique prediction model, resulting in many specialized models without benefitting from the experience embedded in data owned by other clients. Similarly to central learning, our local models are trained for 100 epochs. We consider the early stopping callback based on the lowest loss achieved on the validation set and the model checkpoint callback.

### C. Federated Learning

We apply the general FL architecture (FIGURE 8) and extend it with clustered aggregation (TABLE 2).

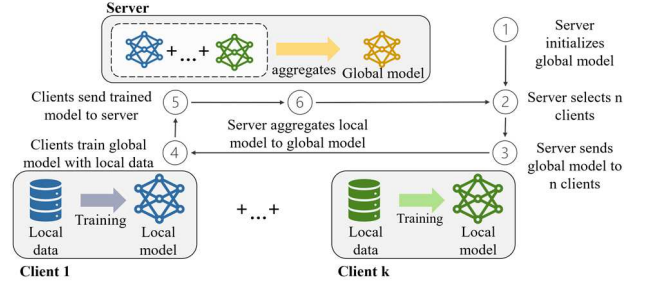

FIGURE 8: FEDERATED LEARNING ARCHITECTURE

First, the server randomly initializes a global model with cluster-specific model weights. Within each cluster, the server distributes the global model to the clients, where each participating client i trains the model on its local data $D_i$. Here we apply the Adam optimizer [35], a version of stochastic gradient descent. Afterward, the clients return their updated model parameters to the server, where the global model is updated using weighted averaging. The server repeats this procedure for t training rounds [21]. To update our clients, we apply the federated averaging algorithm, which calculates the data-weighted average for each model.

TABLE 2: FL ALGORITHM WITH CLUSTERED AGGREGATION

| Algorithm 1: Federated training with clustered aggregation |   |
| --- | --- |
| 1 | Input: Attribute data from each of m clients |
| 2 | Based on attributes, the server groups clients into n clusters |
| 3 | The server randomly initializes a base model $w_{rand}$ |
| 4 | **for** each cluster $C^k$ with k=1,2,…,N, **in parallel do** |
| 5 |    Initialize cluster-specific model weights, $w_k \leftarrow w_{rand}$ |
| 6 | **end for** |
| 7 | **for** each cluster $C^k$ with k=1,2,…,N, **in parallel do** |
| 8 |    **for** communication round t=1,2,…,T **do** |
| 9 |      **for** each client $c^i$ in cluster $C^k$, i=1,2,…,T, **in parallel do** |
| 10 |       Synchronize the local model with the latest cluster specific model: $w^i_k \leftarrow w_k$ |
| 11 |       Update local model $w^i_k$ by training on local Data, $D^i = \{X^i, Y^i\}$ |
| 12 |       Transmit updated $w^i_k$ back to the server |
| 13 |      **end for** |
| 14 |      Update model weights for the cluster-specific model by aggregation: $w_k \leftarrow \frac{1}{l}\sum_{i=1}^{l} w^i_k$ |
| 15 |    **end for** |
| 16 | **end for** |
| 17 | Output: cluster-specific models with trained weights: $w_k$, k=1,2,…,N |

The performance of FL is best when the data is non-independent and identically distributed (non-iid) [36]. However, training a model with FL usually suffers from the non-iid data problem, where clients contain data distributions that are diverse from each other. As a result, the global model has a poor convergence rate and performance. Clustering clients with similar properties and creating individual global models for each cluster is a promising solution to this problem [21]. To deal with data heterogeneity, we apply K-Means clustering (with k = 6) for the 33 selected clients to obtain six clusters. As the time series sequences may vary in timing and speed, we apply dynamic time warping as a clustering metric. We choose the hyperparameters for our FL architecture after careful manual tuning using partial data from the constructed dataset. The most critical parameters are the number of local epochs $N_{epoch}$ and the number of rounds $N_{round}$ [21]. Considering that clients have restricted processing capabilities, we set $N_{epoch} = 20$ and $N_{round} = 2$.

## V. Performance evaluation

To evaluate the performance of the different training scenarios, we report the RMSE, MAE, MAPE, and training time per epoch for each training scenario. The reported metrics are averaged over all clients, clusters, or sequences. Our results show that both FL and local learning achieve high forecasting performance, while FL works especially well with limited data available. Further, our transformer-based model outperforms the LSTM model and CNN in most scenarios while needing only half the training time compared to the LSTM model.

In central learning, the forecasting models can benefit from large volumes of data, while the models might struggle to generalize and capture individual household behaviors. The metrics in TABLE 3 show that the transformer-based model performs best within the central learning architecture in every training scenario based on RMSE, MAE, and MAPE. Here the highest forecasting accuracy for the next 12 and 24 hours is achieved with the combination of calendar and weather features. However, the CNN is the fastest forecasting algorithm, while the LSTM model needs the longest training time per epoch.

TABLE 3: EVALUATION OF CENTRAL LEARNING

| Horizon | Features | Model | RMSE | MAE | MAPE (e+07) | Time (s) |
|---|---|---|---|---|---|---|
| 12 | 5 | CNN | 1.5628 | 1.4014 | 109.90 | **13** |
|  |  | LSTM | 1.5344 | 1.4055 | 114.03 | 508 |
|  |  | Transf. | 1.0160 | 0.8881 | 62.08 | 385 |
|  | 7 | CNN | 1.8131 | 1.6260 | 134.18 | 51 |
|  |  | LSTM | 1.5344 | 1.4055 | 114.03 | 750 |
|  |  | Transf. | **1.0025** | **0.8844** | **61.11** | 320 |
| 24 | 5 | CNN | 1.5100 | 1.3683 | 108.49 | **26** |
|  |  | LSTM | 1.5372 | 1.4079 | 114.29 | 495 |
|  |  | Transf. | 1.0408 | 0.9331 | 64.95 | 513 |
|  | 7 | CNN | 1.4411 | 1.3111 | 101.61 | 32 |
|  |  | LSTM | 1.5372 | 1.4079 | 114.29 | 793 |
|  |  | Transf. | **1.0221** | **0.9060** | **61.63** | 213 |

The forecasting results of FL and local learning show significantly higher forecasting accuracy. For instance, the best-performing model achieves an RMSE of 0.1677 in local learning and 0.1659 in FL, compared to 1.0025 in the central architecture. The difference implies that the central models struggle to generalize and capture individual behaviors in energy usage. More extensive models might allow predicting individual behaviors, but as data is stored in a single location, the privacy risk to energy consumers is the highest in this training approach.

In FL and local learning, the LSTM and transformer models achieve similar forecasting accuracy. However, the transformer model needs less than half of the training time per epoch, resulting in higher efficiency and scalability.

In the local training, each model trains in isolation. This approach allows the model to learn individual load patterns while preventing to benefit from the other datasets. As in the central architecture, the transformer model achieves the highest accuracy in local learning (TABLE 4). However, no definite statement on the best feature set can be made. Both feature sets yield the best results depending on the horizon and evaluation metrics.

TABLE 4: EVALUATION OF LOCAL LEARNING

| Horizon | Features | Model | RMSE | MAE | MAPE (e+07) | Time (s) |
|---|---|---|---|---|---|---|
| 12 | 5 | CNN | 0.2094 | 0.1787 | 11.168 | 0.45 |
|  |  | LSTM | 0.1950 | 0.1672 | 10.459 | 6.66 |
|  |  | Transf. | 0.1680 | 0.1347 | 7.239 | 2.26 |
|  | 7 | CNN | 0.2111 | 0.1802 | 11.693 | **0.38** |
|  |  | LSTM | 0.1992 | 0.1739 | 10.827 | 7.32 |
|  |  | Transf. | **0.1677** | **0.1323** | **6.896** | 2.89 |
| 24 | 5 | CNN | 0.2080 | 0.1786 | 11.197 | **0.45** |
|  |  | LSTM | 0.1998 | 0.1741 | 10.904 | 8.46 |
|  |  | Transf. | **0.1705** | 0.1392 | 7.656 | 2.76 |
|  | 7 | CNN | 0.2095 | 0.1781 | 11.141 | 0.55 |
|  |  | LSTM | 0.1984 | 0.1728 | 10.786 | 9.62 |
|  |  | Transf. | 0.1714 | **0.1384** | **7.482** | 3.63 |

For the FL architecture (TABLE 5), the LSTM and transformer model achieve similar forecasting accuracy, while the transformer needs 48% less training time per epoch. Considering a 12-hour horizon, the transformer-based model yields the lowest RMSE. However, the LSTM model performs best regarding MAE and MAPE. For a forecasting horizon of 24 hours, the LSTM model outperforms the transformer in accuracy.

TABLE 5: EVALUATION OF FEDERATED LEARNING

| Horizon | Features | Model | RMSE | MAE | MAPE (e+07) | Time (s) |
|---|---|---|---|---|---|---|
| 12 | 5 | CNN | 0.2111 | 0.1777 | 10.828 | 0.54 |
|  |  | LSTM | 0.1662 | **0.1309** | 7.0647 | 8.12 |
|  |  | Transf. | **0.1659** | 0.1315 | 7.0747 | 2.58 |
|  | 7 | CNN | 0.2135 | 0.1802 | 11.503 | **0.34** |
|  |  | LSTM | 0.1665 | 0.1313 | **7.0401** | 6.45 |
|  |  | Transf. | 0.1662 | 0.1327 | 7.1314 | 3.07 |
| 24 | 5 | CNN | 0.2122 | 0.1794 | 10.974 | 0.65 |
|  |  | LSTM | **0.1701** | **0.1375** | **7.5577** | 8.77 |
|  |  | Transf. | 0.1711 | 0.1404 | 7.7486 | 5.05 |
|  | 7 | CNN | 0.2094 | 0.1777 | 10.952 | **0.46** |
|  |  | LSTM | 0.1730 | 0.1406 | 7.8197 | 7.89 |
|  |  | Transf. | 0.1713 | 0.1392 | 7.5743 | 3.79 |

The similar forecasting performance of FL and local learning is surprising, as the clustered FL architecture should benefit from similar datasets within the clusters. One hypothesis to explain the resemblance could be the large training dataset. Thus, additional data from different users would be redundant. To evaluate this hypothesis, we perform further experiments only using the first three months of the datasets for training. Here the results show that the FL architecture (TABLE 6) outperforms the local learning (TABLE 7) by 3% accuracy with limited data available, compared to 1% with the large dataset. The increasing difference indicates that FL may be most relevant for STLF with restricted data available. For instance, when advanced metering infrastructure has just been installed. To avoid waiting several months before making reliable predictions, FL can help increase accuracy from the beginning.

TABLE 6: EVALUATION FEDERATED LEARNING WITH 3-MONTH DATA

| Hori-zon | Fea-tures | Model | RMSE | MAE | MAPE (e+07) | Time (s) |
|---|---|---|---|---|---|---|
| 12 | 5 | CNN | 0.2104 | 0.1818 | 11.569 | 3 |
|  |  | LSTM | 0.1808 | 0.1540 | 8.5709 | 46 |
|  |  | Transf. | 0.1768 | 0.1478 | 8.0747 | 24 |
|  | 7 | CNN | 0.2153 | 0.1855 | 12.275 | 3 |
|  |  | LSTM | 0.1876 | 0.1608 | 9.1524 | 46 |
|  |  | Transf. | **0.1736** | **0.1437** | **7.9205** | 25 |
| 24 | 5 | CNN | 0.2059 | 0.1783 | 11.067 | 4 |
|  |  | LSTM | 0.1925 | 0.1681 | 9.6662 | 54 |
|  |  | Transf. | **0.1775** | **0.1498** | **8.3008** | 28 |
|  | 7 | CNN | 0.2057 | 0.1761 | 11.200 | 4 |
|  |  | LSTM | 0.1926 | 0.1678 | 9.6334 | 59 |
|  |  | Transf. | 0.1806 | 0.1537 | 8.4696 | 33 |

TABLE 7: EVALUATION LOCAL LEARNING WITH 3-MONTH DATA

| Hori-zon | Fea-tures | Model | RMSE | MAE | MAPE (e+07) | Time (s) |
|---|---|---|---|---|---|---|
| 12 | 5 | CNN | 0.2124 | 0.1844 | 11.881 | 2 |
|  |  | LSTM | 0.1936 | 0.1684 | 9.4323 | 44 |
|  |  | Transf. | **0.1787** | **0.1497** | **8.1209** | 20 |
|  | 7 | CNN | 0.2222 | 0.1917 | 12.750 | 2 |
|  |  | LSTM | 0.1952 | 0.1705 | 9.6137 | 39 |
|  |  | Transf. | 0.1822 | 0.1547 | 8.4378 | 19 |
| 24 | 5 | CNN | 0.2065 | 0.1796 | 11.271 | 3 |
|  |  | LSTM | 0.1950 | 0.1701 | 9.5905 | 54 |
|  |  | Transf. | 0.1844 | 0.1583 | 8.7302 | 25 |
|  | 7 | CNN | 0.2127 | 0.1845 | 11.914 | 3 |
|  |  | LSTM | 0.1945 | 0.1697 | 9.5587 | 57 |
|  |  | Transf. | **0.1828** | **0.1557** | **8.5105** | 27 |

Based on our findings, we summarize the following recommendations:

- For the forecasting horizons studied (12 and 24 hours), the transformer model achieves one of the highest accuracies (based on RMSE, MAE, and MAPE) with comparatively low training time per epoch. Thus, our transformer model presents an accurate and scalable alternative to LSTM models and CNNs.
- A local or FL architecture should be chosen if data security and prediction accuracy are highly relevant. FL is particularly suitable for use cases with limited data available.
- For STLF, a combination of calendar and weather features can be helpful. However, feature selection should be performed individually depending on the dataset.

## VI. CONCLUSION

In this paper, we presented a novel transformer-based model within an FL architecture for STLF. We benchmarked our FL architecture against central and local learning to evaluate our results and compare our model's performance to LSTM models and CNNs. Our simulations showed that transformer-based forecasting is a promising alternative to state-of-the-art models within local, central, and FL. We determined that FL approaches can outperform centralized learning and slightly outperform local learning. However, we presented favorable results with limited data available. In this case, the performance of the FL model can be improved by 3% compared to local learning while retaining privacy. Further, we demonstrated that the transformer model achieved one of the highest accuracies with 48% less training time than the LSTM model. Future work could study the effect of different clustering techniques and varying cluster sizes on forecasting accuracy.

ACKNOWLEDGMENT

We acknowledge support by the KIT-Publication Fund of the Karlsruhe Institute of Technology. The presented work was funded by the German Research Foundation (DFG) as part of the Research Training Group 2153: 'Energy Status Data—Informatics Methods for its Collection, Analysis, and Exploitation'.